\documentclass{llncs}
\usepackage{times}
\usepackage{um97}
\advance\topmargin by -1 in
\advance\oddsidemargin by -1 in
\evensidemargin=\oddsidemargin
\begin{document}

\title{Empirically Evaluating an Adaptable Spoken Dialogue System}

\author{Diane J. Litman\inst{1} and Shimei Pan\inst{2}\thanks{We
thank J. Chu-Carroll, C. Kamm, D. Lewis,  M. Walker,
and S. Whittaker for helpful comments.}
\institute{AT\&T Labs - Research, Florham Park, NJ, USA
\and
Computer Science Department, Columbia University, New York, NY,  USA}}

\maketitle

\begin{abstract}
Recent technological advances have made it possible to build
real-time, interactive spoken dialogue systems for a wide variety of
applications.  However, when users do not respect the limitations of
such systems, performance typically degrades.  Although users differ
with respect to their knowledge of system limitations, and although
different dialogue strategies make system limitations more apparent to
users, most current systems do not try to improve performance by
adapting dialogue behavior to individual users.
This paper presents an empirical evaluation of TOOT, an adaptable
spoken dialogue system for retrieving train schedules on the web.  We
conduct an experiment in which 20 users carry out 4 tasks
with both adaptable and non-adaptable versions of TOOT,
resulting in a corpus of 80 dialogues.  The values for a wide range of
evaluation measures are then extracted from this corpus.  Our results
show that adaptable TOOT generally outperforms non-adaptable TOOT, and that
the utility of adaptation depends on TOOT's initial dialogue
strategies.
\end{abstract}

\section{Introduction}

Recent technological advances have made it possible to build
real-time, interactive spoken dialogue systems for a wide variety of
applications.  
However, due to limitations in such areas as
vocabulary, grammar, and application functionality, most deployed
systems can only accept a small set of natural language inputs from
users.  When users do not respect the limitations of spoken dialogue
systems, system performance typically degrades.  
For example, user utterances are more likely to be rejected or misunderstood,
dialogue length typically increases, and users are
less likely to achieve their task goals.
Although individual users differ with respect to their knowledge of
system limitations~(\withincite{KLW98}), and although different dialogue
strategies make system limitations more apparent to
users~(\withincite{WHFDM97};~\withincite{DanieliGerbino95}), current spoken dialogue systems
typically do not try to adapt their dialogue strategies to individual users.
A major obstacle to having a spoken dialogue system perform such
adaptation dynamically is that there is currently little understanding
of the type of properties that would need to be represented in a user
model for this task, let alone automatically acquired during the
course of a dialogue.  Nevertheless,
users do seem to intuitively know when and how their interactions with
a spoken dialogue system could be changed, in order to improve overall
system performance.  This observation has led us to explore the
utility of an {\it adaptable} spoken dialogue system, where the {\it
user} rather than the system controls the adaptation of the system's
dialogue strategies.  In particular, we hypothesize that if a user can
effectively adapt the system's behavior, 
so that the system will use the dialogue strategies that
best match the user's needs at any point in a dialogue,
then the overall performance of the system should improve.

Figure~\ref{TOOT} shows an example dialogue with an adaptable train-timetable spoken dialogue system
called TOOT.\footnote{This excerpt is taken from the experimental corpus
described below.}
\begin{figure}
\small{
\begin{tabular}{ll} 
U1: &I'd like to go to Chicago from Baltimore on Saturday at 8 am.\\
T1: &Do you want to leave in the morning or in the evening?\\
U2: &Uh morning.\\
T2: &Sorry, I can't understand you.  Please repeat your answer.\\
U3: &In the morning.\\
T3: &Do you want me to find the trains from Baltimore to Chicago on Saturday around 10 morning now?\\
U4: &No.\\
T4: &How may I help you?  \\
U5: &I would like to go from Baltimore to Chicago on Saturday at 8 o'clock.\\
T5: &Do you want me to find the trains from Baltimore to DC on Tuesday around 11:45 am now?\\
U6: &No.\\
T6: &How may I help you?\\
U7: &Change strategy.	\\
T7: &You are using the user initiative.  Which initiative do you want to change to? System --\\
U8: &System initiative please. \\
T8: &You are using the no confirmation strategy.  Which confirmation strategy do you want to change to?\\
   & No confirmation, implicit confirmation, or explicit confirmation?\\
U9: &Implicit please.\\
T9: &I heard you say go to DC on Tuesday around 11:45 am.\\
   & Do you want me to find the trains from Baltimore to DC on Tuesday around 11:45 am now?\\
U10: &No.\\
T10: &Which city do you want to go to?\\
U11: &Chicago.\\
T11: &I heard you say go to Chicago. Which city do you want to leave from?\\
U12: &Baltimore.\\
T12: &I heard you say leave from Baltimore. On which day of the week do you want to leave?\\
U13: &Saturday.\\
T13: &I heard you say leave on Saturday. What time do you want to leave?\\
U14: &8:00 in the morning.\\
T14: &Sorry, I can't understand you.  Please repeat your answer.\\
U15: &8:00 in the morning.\\
T15: &I heard you say leave around 8:00 in the morning. Do you want me \\
     &to find the trains from Baltimore to Chicago on Saturday around 8:00 in the morning now?\\
U16: &Yes.
\end{tabular}
\caption{Example dialogue with an adaptable spoken dialogue system.}
\label{TOOT}}
\end{figure}
In this excerpt, TOOT initially uses a set of dialogue strategies
often found in human-human interactions.  However, the strategies do
not work particularly well for this user. Utterances U1 and U5
are misunderstood by TOOT, and U2 is not understood at all.
Recognizing a mismatch, the user changes the strategies used
by TOOT (U7 - U9).  
As a result, the open ended prompt that TOOT used
earlier in the dialogue (T4) is now replaced with a series of specific
questions (T10, T11, T12, and T13).
TOOT also now
highlights any potential misrecognitions by
prefacing its utterances with ``I heard you say \ldots''.  Earlier in
the dialogue, it took 5 utterances before the user was aware that TOOT had 
misrecognized the ``8'' in U1 as ``10''.
As a result of these changes, the dialogue proceeds more smoothly after
the adaptation (e.g., TOOT's misrecognition rate is reduced), and a correct
database query is soon generated (T15-U16).

In this paper, we present an  evaluation of
adaptability in TOOT.
We conduct an experiment in which 20 novice
users carry out 4 tasks
with one of two versions
of TOOT ({\it adaptable} and {\it non-adaptable}), resulting in a
corpus of 80 dialogues.  The values for a range of evaluation
measures are then extracted from this corpus.  
Hypothesis testing shows that a
variety of differences depend on the user's ability to adapt the
system.  A PARADISE assessment of the contribution of each evaluation
measure to overall performance~(\withincite{Walkeretal97}) shows that the phenomena influenced by
adaptation are also the major phenomena that significantly influence
performance.  Our results show that adaptable TOOT generally outperforms
non-adaptable TOOT, and that the utility of adaptation 
depends on the initial configuration of dialogue strategies.

\section{TOOT}

TOOT is a voice-enabled dialogue system for accessing train schedules
from the web via a telephone conversation. TOOT is implemented using a
spoken dialogue system platform~(\withincite{Kammetal97}) that combines
automatic speech recognition (ASR), text-to-speech (TTS), a phone
interface, and modules for specifying a dialogue manager and
application functions.  ASR in our platform is speaker-independent,
grammar-based and supports {\it barge-in} (which allows users to
interrupt TOOT when it is speaking, as in utterances T7 and U8 in
Figure~\ref{TOOT}).  The dialogue manager uses a finite state machine
to control the interaction, based on the current system state and ASR
results.  TOOT's application functions access train schedules
available at www.amtrak.com.  Given a set of constraints, the
functions return a table listing all matching trains in a specified
temporal interval, or within an hour of a specified timepoint. This
table is converted to a natural language response which can be
realized by TTS through the use of templates.\footnote{The current version of TOOT uses
a literal response strategy~(\withincite{LPW98}).  Informally, if the returned table contains 1-3 trains, TOOT lists
the trains; if the table contains greater than 4 trains, TOOT lists the trains 3 at a time; if the table is empty, TOOT reports that no trains
satisfy the constraints.  
TOOT then asks the user if she wants to continue and find a new set of trains.}

Depending on the user's needs during the dialogue, TOOT can use one of
three dialogue strategies for managing initiative (``system'',
``mixed'' or ``user''), and one of three strategies for managing
confirmation (``explicit,'' ``implicit,'' or ``no'').
TOOT's initiative
strategy specifies who has control of the dialogue, while TOOT's
confirmation strategy specifies how and whether TOOT lets the user
know what it just understood. In Figure~\ref{TOOT}, TOOT initially
used user initiative and no confirmation, then later used system
initiative and implicit confirmation.  The following fragments provide
additional illustrations of how dialogues vary with strategy:
\begin{tabular}{llll}
	&&&\\
     &{\it System Initiative, Explicit Confirmation} &&{\it User Initiative, No Confirmation}\\
  T: &Which city do you want to go to?  &T: &How may I help you? \\
  U: &Chicago.			  	&U: &I want to go to Chicago from Baltimore.\\
  T: &Do you want to go to Chicago?	&T: &On which day of the week do you want to leave?\\
  U: &Yes.	     			&U: &I want a train at 8:00.\\
	 &			&&\\	
\end{tabular}
Although system initiative with explicit confirmation is the most
cumbersome approach, it can help improve some aspects of performance for users who do not have a
good understanding of the system's limitations.  The use of system
initiative helps reduce ASR misrecognitions and
rejections~(\withincite{WHFDM97}), by helping to keep the user's utterances within the
system's vocabulary and grammar.  The use of explicit confirmation
helps increase the user's task success~(\withincite{DanieliGerbino95}), 
by making the user more aware of any ASR misrecognitions and making
it easier for users to correct misrecognitions when they occur.
On the other hand, system initiative and explicit confirmation typically
increase total dialogue 
length~(\withincite{WHFDM97};~\withincite{DanieliGerbino95}).  For users whose utterances
are generally understood, other strategies might be more effective.
Consider the use of user initiative with no confirmation, the most
human-like approach. In user (as well as in mixed) initiative mode,
TOOT can still ask the user specific questions, but can also ask open-ended questions such as ``How may I help you?''.  Furthermore, in user (but not in
mixed) initiative mode, TOOT even lets the user ignore TOOT's
questions (as in the last user utterance in the example above). 
By allowing users to specify multiple
attributes in a single utterance, and by not informing users of every
potential misrecognition, this approach can lead to very short
dialogues when ASR performance is not a problem.

In an earlier implementation of TOOT
(as well as in other spoken dialogue systems that we have studied~(\withincite{WHFDM97};~\withincite{KLW98})),
a set of initial dialogue strategies was assigned to the system as a default for
each user, and could not be changed if inappropriate.\footnote{In particular, the previous
version of TOOT always used system initiative with implicit confirmation~(\withincite{LPW98}).}    As discussed
above, however, we hypothesize that we can improve TOOT's
performance by dynamically adapting the choice of dialogue
strategies, based on the circumstances at hand.
Although one of our long-term goals is to have TOOT automatically control the adaptation process, 
this would require that we first solve several open research topics.
For example, TOOT would need to be able to detect, in real time,
dialogue situations suggesting
system adaptation.
As a result, our initial research has instead focused on
giving users the ability to dynamically adapt TOOT's dialogue behaviors.
For
example, if a user's utterances are not being understood, the user
could try to reduce the number of ASR rejections and misrecognitions
by changing the strategies so that TOOT would take more
initiative.  Conversely, if a user's utterances are  being correctly
understood, the user could try to decrease the dialogue length by
having TOOT perform less confirmations.  To allow us to test
whether such an adaptable system does indeed increase performance, we
have created both ``adaptable'' and ``non-adaptable'' versions of
TOOT.  In adaptable TOOT, users are allowed to say ``change strategy''
at any point(s) in the dialogue.  TOOT then asks the user to specify
new initiative and confirmation strategies, as
in utterances U7-U9 in Figure~\ref{TOOT}.  In
non-adaptable TOOT, the default dialogue strategies can not be
changed.

\section{Experimental Design}

Our experiment was designed to test if adaptable TOOT performed
better than non-adaptable TOOT, and whether any differences 
depended on TOOT's initial dialogue strategies and/or the
user's task.  Our design thus consisted of three factors:
{\it adaptability}, {\it initial dialogue strategy},
and {\it task scenario}.  Subjects were 20 AT\&T technical summer
employees not involved with the design or implementation of TOOT, 
who were also novice users of spoken dialogue systems in general.  10 users
were randomly assigned to {\it adaptable} TOOT and 10 to {\it
non-adaptable} TOOT.  For each of these groups, 5 users were randomly
assigned to a version of TOOT with the initial dialogue strategies set
to {\it system initiative} and {\it explicit confirmation} (SystemExplicit TOOT);
the remaining 5 users were assigned to a version of TOOT with the
initial dialogue strategies set to {\it user initiative} and
{\it no confirmation} (UserNo TOOT).  
Each user performed the same  4 tasks in sequence.  
Our experiment yielded a corpus of 80
dialogues (2633 turns; 5.4 hours of speech).

Users used the web to read a set of experimental instructions in their
office, then called TOOT from their phone.  The experimental
instructions consisted of a description of TOOT's functionality, hints
for talking to TOOT, and links to 4 task scenarios.  
An example task scenario is as follows: ``Try to find a train going
{\it to} {\bf Chicago} {\it from} {\bf Baltimore} on {\bf Saturday} at
{\bf 8 o'clock am}. If you cannot find an exact match, find the one
with the {\bf closest} departure time.  Please write down the {\bf
exact departure time} of the train you found as well as the {\bf total
travel time}.''  
The instructions for
adaptable TOOT also contained a brief tutorial explaining how to
use ``change strategy'', and guidelines for doing so (e.g., `` if you
don't know what to do or say, try system initiative'').  

We collected three types of data to compute a number of measures relevant for spoken
dialogue evaluation~(\withincite{WHFDM97}). 
First, all dialogues were recorded. 
The recordings were used to calculate the total time of each dialogue (the evaluation
measure {\bf Elapsed Time}), and
to (manually) count how
many times per dialogue each user interrupted TOOT ({\bf Barge Ins}).

Second, the dialogue manager's behavior
on entering and exiting
each state in the finite state machine was logged.
This log was used to calculate the total number of
{\bf System Turns} and {\bf User Turns},
{\bf Timeouts} (when the user doesn't say anything within a specified time frame,
TOOT provides suggestions about what to say ),
{\bf Helps } (when the user says ``help'', TOOT provides a context-sensitive help message),
{\bf Cancels } (when the user says ``cancel'', TOOT undoes its previous action), and
{\bf ASR Rejections} (when the confidence level of ASR is too low, TOOT asks the user to repeat
the utterance).
In addition, by listening to the recordings and comparing them
to the logged ASR results, we calculated the concept accuracy
(intuitively, semantic interpretation accuracy) 
for each utterance.
This was then used, in combination with ASR rejections, to compute
a {\bf Mean Recognition} score per dialogue.

Third, users filled out a web survey after each dialogue.
Users specified the departure and travel times that they obtained via
the dialogue.  Given that there was a single correct train to be
retrieved for each task scenario, this allowed us to determine whether
users successfully achieved their task goal or not ({\bf Task Success}).
Users also responded to the following questionnaire:
\begin{itemize}
\item Was the system easy to understand?  ({\bf TTS Performance})
\item Did  the system  understand what you said? ({\bf ASR Performance})
\item Was it easy to find the schedule you wanted? ({\bf Task Ease})
\item Was the pace of interaction with the system appropriate? ({\bf Interaction Pace})
\item Did you know what you could say at each point of the dialogue? ({\bf User Expertise})
\item  How often was the system sluggish and slow to reply to you? ({\bf System Response})
\item  Did the system  work the way you expected it to? ({\bf Expected Behavior})
\item From your current experience with using our system,
do you think you'd  use this  regularly to access
train schedules when you are away from your desk? ({\bf Future Use})
\end{itemize}
Each question 
measured a particular usability
factor, e.g., {\bf TTS Performance}.  Responses ranged over {\it n}
pre-defined values (e.g., {\it almost never, rarely, sometimes, often,
almost always}), and were mapped to an integer in $1 \ldots 5$ (with
5 representing optimal performance).
{\bf User Satisfaction} was computed by summing each
question's score, and thus ranged in value from 8 to 40.

\section{Results}

We use analysis of variance (ANOVA)~(\withincite{Cohen95}) to determine
whether the adaptability of TOOT produces significant
differences in any of the evaluation measures for our experiment.  
We also use the PARADISE evaluation framework~(\withincite{Walkeretal97}) to
understand which of our evaluation measures best predicts overall performance in TOOT. Following PARADISE, we organize
our evaluation measures along the following four performance
dimensions:
\begin{itemize}
\item {\it task success}: Task Success 
\item {\it dialogue quality}: Helps, ASR Rejections, Timeouts, Mean Recognition, Barge Ins, Cancels
\item {\it dialogue efficiency}: System Turns, User Turns, Elapsed Time
\item {\it system usability}: User Satisfaction (based on TTS Performance, ASR Performance, Task Ease, Interaction Pace, User Expertise, System Response, Expected Behavior, Future Use)
\end{itemize}

\subsection{Adaptability Effects}

Recall that our mixed\footnote{{\it Task scenario} is between-groups and {\it initial dialogue strategy} and {\it adaptability} are within-group.} experimental design consisted of  three factors:
{\it adaptability}, {\it initial dialogue strategy}, and {\it task scenario}.
Each of our evaluation measures is analyzed using a three-way ANOVA 
for these three factors.
The ANOVAs demonstrate a {\it main effect of  adaptability} for the 
task success and 
system usability dimensions of performance.  
These main adaptability effects are independent of TOOT's initial
dialogue strategy as well as of the task scenario being executed by the user.
The ANOVAs also demonstrate
{\it interaction effects of  adaptability and initial dialogue strategy} for
the dialogue quality and system usability performance
dimensions.
In contrast to the main effects, these adaptability effects are not
independent of TOOT's initial dialogue strategy (i.e., the effects of
adaptability and initial strategy are not additive).\footnote{Effects of
{\it initial dialogue strategy} and {\it task scenario} are beyond the scope of this paper.}

Table~\ref{main} summarizes the means for
each evaluation measure that shows a main effect of adaptability, and that cannot
be further explained by any interaction effects.
The first row in the table indicates that 
Task Success is significantly higher for adaptable TOOT
than for non-adaptable TOOT.
Users successfully achieve the goals specified in the task scenario
in 80\% of the dialogues with adaptable TOOT, 
but in only 55\% of the dialogues with non-adaptable TOOT.  
The probability p$<$.03 indicates that the difference is statistically significance
(the standard upper bound for calling a result statistically significant is p$<$.05~(\withincite{Cohen95})).
The second row indicates that with respect to
User Satisfaction, users also rate adaptable TOOT more highly than
non-adaptable TOOT.  
Recall that User Satisfaction takes all of the factors in the usability
questionnaire into account. As will be discussed below, PARADISE correlates
overall system performance with this measure.
In sum, our ANOVAs indicate that making TOOT adaptable increases users' rates of task success as well as users' perceptions of overall system usability.
\begin{table}[t]
\small{
\begin{center}
\caption{Main effects of adaptability.}
\label{main}
\begin{tabular}{|l|r|r|} \hline
Measure					&Non-Adaptable (n=40)	&Adaptable (n=40) 	\\ \hline 
Task Success (\%) (p$<$.03)			&55.00			&80.00			\\ 
User Satisfaction (p$<$.03)	&26.68			&31.60			\\ \hline
\end{tabular}
\end{center}
\begin{center}
\caption{Interaction effects of adaptability and initial dialogue strategy.}
\label{interaction}
\begin{tabular}{|l|r|r|r|r|} \hline
&\multicolumn{2}{|c|}{Non-Adaptable (n=40)} & \multicolumn{2}{|c|}{Adaptable (n=40)}\\ \hline
Measure					&SystemExplicit	&UserNo &SystemExplicit	&UserNo 	\\ \hline 
Mean Recognition (\%) (p$<$.01)  	&88.44		&57.94	&82.55		&75.85	\\ 
User Expertise (p$<$.05)		&4.69		&3.01	&4.45		&3.85	\\ 
Future Use (p$<$.02)			&3.50		&1.70	&3.60		&3.80	\\ \hline
\end{tabular}
\end{center}
}
\end{table}

Table~\ref{interaction} summarizes the means for each evaluation measure
that shows an interaction effect of adaptability and initial dialogue strategy. 
A similar pattern of interaction emerges in the first and second rows
of the table. When users are given the capability to adapt TOOT, Mean
Recognition decreases for SystemExplicit TOOT (88.44\% versus 82.55\%)
but increases for UserNo TOOT (57.94\% versus 75.85\%).  Perceptions
of User Expertise also decrease for adaptable SystemExplicit TOOT
(4.69 versus 4.45) but increase for adaptable UserNo TOOT (3.01 versus
3.85).  In contrast, Future Use is higher for adaptable TOOT than for
non-adaptable TOOT, for both initial strategies.  Thus, users of
adaptable TOOT are more likely than users of non-adaptable TOOT to
think that they would use TOOT on a regular basis.  However, the
increase in Future Use is smaller for SystemExplicit TOOT (3.5 to 3.6)
than for UserNo TOOT (1.7 to 3.8).  In sum, Table~\ref{interaction}
indicates that differences reflecting both dialogue quality and system
usability are an effect of the interaction of the adaptability of TOOT
and TOOT's initial dialogue strategy.  For the UserNo version of TOOT,
making TOOT adaptable increases the means for all of the measures
shown in Table~\ref{interaction}.  For the SystemExplicit version of
TOOT, despite the Mean Recognition and User Expertise results
in Table~\ref{interaction}, users are nevertheless at least if not more likely to
use adaptable System Explicit TOOT in the future.
We speculate that users are willing to tolerate minor levels of
particular types of performance degradations in SystemExplicit TOOT, in order to
obtain the sense of control provided by adaptability.
We also speculate that the utility of adaptable SystemExplicit TOOT would increase
for expert users.
In conjunction with Table~\ref{main}, our results with novice users
suggest that adaptability
is an extremely useful capability to add to UserNo TOOT, and a
capability that is still worth adding to SystemExplicit TOOT.

It is interesting to also examine the way in which adaptation is performed
for each initial dialogue strategy.  Of the 20 dialogues with adaptable
SystemExplicit TOOT, 5 dialogues contained 1 adaptation and a 6th
dialogue contained 2 adaptations.  3 of the 5 users adapted at least
1 dialogue, and overall, confirmation was changed more times than
initiative.
Of the 20 dialogues with adaptable UserNo TOOT, 10
dialogues contained 1 adaptation and an 11th dialogue contained 2
adaptations. All 5 users of UserNo TOOT adapted at least 1 dialogue.
Users of UserNo TOOT changed initiative more than they changed
confirmation, and also changed initiative more drastically.
In conjunction with
our ANOVA results, these observations lead us to speculate that
adapting   a poorly performing system is both more feasible and more
important for novice users than adapting a reasonably performing system. 

\subsection{Contributors to Performance}

To quantify the relative importance of our multiple evaluation
measures to performance, we use the PARADISE evaluation framework to
derive a performance function from our data.  The
PARADISE model posits that performance can be correlated with a
meaningful external criterion of usability such as User
Satisfaction.  PARADISE then uses stepwise multiple linear regression to model
User Satisfaction from measures representing the performance dimensions of task success, dialogue
quality, and dialogue efficiency:
\[\mbox{User  Satisfaction} = \sum_{i=1}^{n} w_{i} \ast {\cal N}(measure_{i})
\]
Linear regression produces coefficients (i.e., weights $w_{i}$) describing the relative
contribution of predictor factors in accounting for the variance in a
predicted factor.
In PARADISE, the task success and dialogue cost measures
are predictors, while User Satisfaction is predicted.
The normalization function $\cal N$ guarantees that
the coefficients directly indicate the relative contributions.

The application of PARADISE to the TOOT data shows that the most
significant contributors to User Satisfaction are
Mean Recognition, Task Success, and 
Elapsed Time, respectively.  In addition, PARADISE shows that the
following performance function provides the best fit to our data,
accounting for 55\% of the variance in User Satisfaction:\footnote{Linear regression
assumes that predictors are not highly correlated (e.g.,
because correlations above .70 can affect the 
coefficients, deletion of redundant predictors is advised~(\withincite{lm})).
There is only 1 positive correlation among our predictors
(between Mean Recognition and Task Success), and it
is well below .70.}
\[\mbox{User Satisfaction}= .45{\cal N}\mbox{(Mean Recognition)}+.33 {\cal N}\mbox{(Task Success)}-.14 {\cal N}\mbox{(Elapsed Time)}\]
Our performance function demonstrates that TOOT performance (estimated
using subjective usability ratings) can be best predicted using a
weighted combination of objective measures of dialogue quality, task
success, and dialogue efficiency.  In particular, more accurate speech
recognition, more success in achieving task goals, and shorter
dialogues all contribute to increasing perceived performance in TOOT.

Our performance equation helps explain the main effect of adaptability
for User Satisfaction that was shown in Table~\ref{main}.
Recall that our ANOVAs for both Mean Recognition and Task Success
showed adaptability effects (Tables~\ref{interaction} and~\ref{main},
respectively).  Our PARADISE analysis showed that these measures
were also the most important measures in explaining the variance in
User Satisfaction. It is thus not surprising that User
Satisfaction shows an effect of adaptability, with users
rating the performance of adaptable TOOT more highly than non-adaptable TOOT.

A result that was not apparent from the analysis of variance is that
Elapsed Time is a performance predictor.  However, the weighting of
the measures in our performance function suggests that Mean
Recognition and Task Success are more important measures of overall
performance than Elapsed Time.  These findings are consistent with our
previous PARADISE evaluations, where measures of task success and
dialogue quality were also the most important performance
predictors~(\withincite{LPW98};~\withincite{WFN98};~\withincite{KLW98}). Our findings draw into question a
frequently made assumption in the field regarding the centrality of
efficiency to performance, and like other recent work,
demonstrates that there are important tradeoffs between efficiency and 
other performance dimensions~(\withincite{DanieliGerbino95};~\withincite{WHFDM97}).

\section{Related Work}

In the area of spoken dialogue, \cite{vanZanten} has proposed a method for
adapting initiative in form-filling dialogues.
Whenever the system rejects a user's utterance, the system takes more
initiative; whenever the user gives an over-informative answer, the
system yields some initiative.  While this method has the potential
of being automated, the method has been neither fully implemented
nor empirically evaluated.
\cite{Smith98} has evaluated
strategies for dynamically deciding whether to confirm
each user utterance during a task-oriented dialogue.  
Simulation results suggest that context-dependent adaptation strategies can improve performance,
especially when the system has greater initiative.
\cite{WFN98} and~\cite{LP97} have used reinforcement learning
to adapt dialogue behavior over time such that system
performance improves.  We have instead focused on optimizing performance
during a single dialogue.

The empirical
evaluation of an adaptive interface in a commercial software
system~(\withincite{Strachan97}) is also similar to our work.
Analysis of variance
demonstrated that an adaptive interface based
on minimal user modeling
improved subjective user satisfaction ratings.

\section{Conclusion}

We have presented an empirical evaluation of adaptability in TOOT, a
spoken dialogue system that retrieves train schedules from the web.
Our results suggest that adaptable TOOT generally outperforms
non-adaptable TOOT for novice users, and that the utility of adaptation is greater for
UserNo TOOT than for SystemExplicit TOOT.  By using analysis of
variance to examine how a set of evaluation measures differ as a
function of adaptability, we elaborate the conditions under which
adaptability leads to greater performance.  When users interact with
adaptable rather than non-adaptable TOOT, User Satisfaction and Task Success
are significantly higher.  These results are
independent of TOOT's initial dialogue strategy and task
scenario.  In contrast, Mean Recognition, User
Expertise, and Future Use illustrate an interaction between 
initial dialogue strategy and adaptability.  For SystemExplicit
TOOT, the adaptable version does not outperform the
non-adaptable version, or does not outperform the non-adaptable
version very strongly. For UserNo TOOT, the adaptable version
outperforms the non-adaptable version on all three measures.

By using PARADISE to derive a performance function from 
data, we show that Mean Recognition, Task Success, and Elapsed Time
best predict a user's overall satisfaction with TOOT.  These results
help explain why adaptability in TOOT leads to overall greater
performance, and allow us to make predictions about future
performance.  For example, we predict that a SystemImplicit strategy
is likely to outperform our SystemExplicit strategy, since we expect
that Mean Recognition and Task Success will remain constant but that
Elapsed Time will decrease.

Currently, we are extending our results along two
dimensions.  First, 
we have made a first step towards
automating the adaptation process in TOOT, by using
machine learning to develop a classifier for detecting dialogues
with poor speech recognition~(\withincite{LWK99}). (Recall 
that our PARADISE evaluation suggested that 
recognition accuracy was our best performance predictor.)
We hope to use this classifier to determine the
features that need to be represented in a user model, and
to tell us when the user model indicates the need for adaptation.
Guided by our empirical results, we can then develop an initial
adaptation algorithm that takes dialogue strategy into account.
For example, 
based on our experiment, we would like UserNo TOOT
to adapt itself fairly aggressively when it recognizes
that the user is having a problem.
Second, the experiments reported here considered only our two most
extreme initial dialogue strategy configurations.  To generalize our
results, we are currently experimenting with other dialogue
strategies.  To date we have collected 40 dialogues using a mixed
initiative, implicit confirmation version of TOOT, with initial promising
results.  For example, 
user satisfaction continues to exhibit the same main effect of
adaptability when our corpus is augmented with these new dialogues.

\end{document}